# A Weeding Robot for Seedling Removal

Jarkko Kotaniemi
Intelligent Robotics
VTT Technical Research Centre of Finland Ltd
Oulu, Finland
jarkko.kotaniemi@vtt.fi

Niko Känsäkoski
Intelligent Robotics
VTT Technical Research Centre of Finland Ltd
Oulu, Finland
niko.kansakoski@vtt.fi

Tapio Heikkilä
Intelligent Robotics
VTT Technical Research Centre of Finland Ltd
Oulu, Finland
tapio.heikkila@vtt.fi

*Abstract*— **Automatic weeding technologies have attained a lot of attention lately, because of the harms and challenges weeds are causing for livestock farming, in addition to that weeds reduce yields. We are targeting automatic and mechanical Rumex weeding in open pasture fields using light weight mobile field robot technologies. We describe a mobile weeding robot with GNSS navigation, 3D computer vision for weed detection, and a robot arm with a mechanical weeding tool. Our main contribution is showing the feasibility of light weight robot, sensor, and tool technologies in mechanical removal of weed seedlings.**

## I. Introduction

Automatic weeding technologies have attained a lot of attention lately, because of the harms and challenges weeds are causing for livestock farming. In addition to weeds reducing yields [1], they can be harmful especially for cattle breeding and dairy farming; for example, Longleaf Dock (Rumex longifolius) has high content of oxalates, which at worst, can lead to oxalate poisoning with severe pathological consequences, especially to sheep and horses [2]. In hard weather conditions, like Nordic winters, damages may take place as spots with grass disappeared, leaving place potential for weeds to grow. Current weeding practices mean spraying large amounts of weed herbicides over crop fields that have been genetically adjusted to resist the chemicals [7]. The pesticide and seed industries are huge, according to [7], worth $100 billion globally, and of that, herbicide sales alone account for $26 billion. Robotic weed removal systems with AI are expected to reduce the need of herbicides and genetically modified crops, contributing to green transition in farming.

Rumex weeding can be carried out in many ways. Though treating with herbicides is easiest, many herbicides pose harm to the environment or human beings. Weed removal can also be done mechanically by removing the plant from the ground, or by destroying the root of the plant. Mechanically cutting the stems or removing the whole plant by pulling up by the root is preferrable. For Rumex, this should at best be done while in the seedling stage, when then the roots have not yet developed into a strong tap root. Because of manual weeding being laborious, and should be done frequently, an autonomous weed removal robot is a potential solution to achieve automatic mechanical weed removal [3].

We are targeting automatic and mechanical Rumex weeding in open pasture fields using light weight mobile field robot technologies, developed in the FlexiGrobots project (flexigrobots-h2020.eu). FlexiGrobots introduces a set of autonomous field services for farming, where drones are used to geolocate weeds, resulting to a weed map, which is provided to mobile weeding robots for automatic weeding. Mechanical robotic weeding implies further the robot going to weed locations, local weed detection and finally rooting the weed out mechanically. Earlier we have reported on the detection and localizing of Rumex weed seedlings using 2D and 3D computer vision [9], and here we are taking the further steps to complete weeding missions by mechanical weeding with a mobile robot platform equipped with a robot arm and a special mechanical weeding tool. Our main contribution is showing the feasibility of light weight robot, sensor, and tool technologies in mechanical removal of weed seedlings.

## II. Related work

Weed detection is a critical part of automatic weeding in open pasture fields. Often weed detection is used to detect small weeds growing next to crops or grass. In our case, it is important to detect and classify both the grass, and the plant that needs to be weeded. Basically, weed detection is an image-based classification problem, where weeds are detected from their surroundings as individual plants – here a set of seedling leaves and stems among grass in pastures. Lately especially deep learning methods have been used, like in [4] and [5] for plant disease diagnosis, with a high recognition rate (over 99 %), and in biodiversity monitoring [6]. High reliability rates can be explained by the use of smart phones and high number of persons for data capture, in acquiring large learning data sets, even by non-professionals.

Some developments have been reported, contributing to automatic and mechanical weed removal. A robot system has been proposed by [10], relying on use of a drill to destroy the root of the weed. A robot with a vision system has been used also for precise spraying of herbicides only over the weeds [11]. Plant leaf detection has also been used for transplanting seedlings [12-13] where the importance is in ensuring the integrity of the plant.

There are already some weeding robots in the markets. Odd.Bot [8, 14] has developed a - precision weeding robot for high density crops. It carries out both precise spreading of herbicides and mechanical in-row weed removal, the latter one with a delta robot structure [8, 14]. Pixel Farm Robotics has developed an agricultural robot for smart farming, for cropping, reduced tillage, and smart crop rotation, and also a sensor system equipped with many 3D cameras to be trained to recognize individual plants within a vegetation. However, the robots by Odd.Bot and Pixel Farm Robotics operate in particular type of fields, such as potato fields and "pixel farms", and have not been applied in open pasture fields. An integrated robot system, suitable for open pastures, and where weed detection and removal technologies were integrated, seems still to be missing.

## III. Multilayered Control of Weeding Missions

We have developed a mobile weeding robot with multilayered control. The goal for weed removal is given as a weeding mission, which is set by a weed map of weed locations in the pasture field. The mission consists of tasks for moving the platform, detecting weeds, and performing the weeding action for all the weeds in sequence listed in the weed map. Each task is composed in a fixed manner of primitive actions like robot arm motions, image acquisition, and navigation actions. This multilayered control principle is illustrated in figure 1.

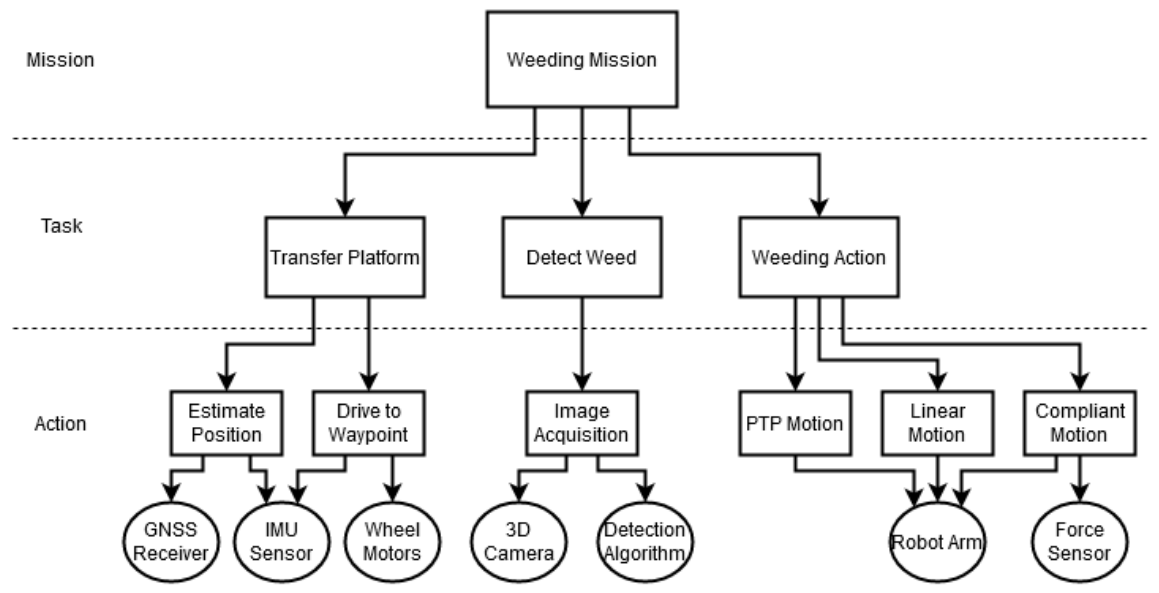

Figure 1. Multilayered control: Missions, tasks, and actions.

For each weed in the weed map there are three tasks to complete. First is to navigate the platform to the weed. This is performed by the transfer platform task, which in turn is made up of actual driving commands to the wheels and estimating the current position and comparing it to the next waypoint in the navigation path. Secondly, when the platform has reached the weed coordinates, the image acquisition task is carried out. In this task an image is taken with the 3D camera and the position of the weed is calculated. Finally, the weeding action -task is executed based on the weed position.

## IV. Weeding Robot Hardware

The platform chassis of the weeding robot from Probot Oy [15] is made from steel (see fig 2). It has two motors in the front for driving the front wheels and two caster wheels in the back for balance. Inside the frame is the battery system with two 48 V 16 Ah batteries for electricity and an amplifier as well as an analogue input module for the force sensor. On top there is a robot arm and emergency switches for the motors and the arm separately.

The hardware architecture of the weeding robot is given in fig 3. The robot is controlled by a laptop computer. There is an IMU sensor for angular velocity and compass measurements. There are two CAN bus converters, one for the motors and one for the arm and force sensor. Attached to the arm there are a force sensor, a 3D camera, and the weeding tool. The weeding tool has three serrated, stainless-steel claws to firmly grab weeds and roots.

The robot arm is a Schunk LWA4P, which is a lightweight 6-DoF robotic arm consisting of three major compact joints called ERB PowerBall modules that integrate two perpendicular axes along with their overall electrical control circuits. Communication to the modules is done by CAN bus with the CANopen standard messages. The robot weighs 12.5 kg and has a maximum workpiece load of 6 kg including the gripper. Each axis has a maximum angular velocity of 72 °/s and the maximum angular acceleration of 150 °/s$^2$. All the axes have the possible angle of rotation of +/- 170° except the third axis which is limited by software to +/- 156.5°. The nominal voltage of the robot is 24 VDC, the average current input is 3 A, and the maximum current input is 12 A.

The force sensor is ME-Systeme K6D40. It is a 6-axis force-torque sensor meaning it can sense linear forces and torques in the 3 different axes. It senses these values by having six embedded strain gauges set in a specific geometric arrangement inside the device. The actual force and torque values are then given by the cross product of a calibration matrix and a vector of the six sensor signals. The calibration matrix is unique for each force sensor and is provided by the manufacturer. The sensor can measure nominally 200 N in the x- and y-axes and 500 N in the z-axis. Torque measurements are max 10 Nm in every axis. The accuracy of the measurements is 0.2%.

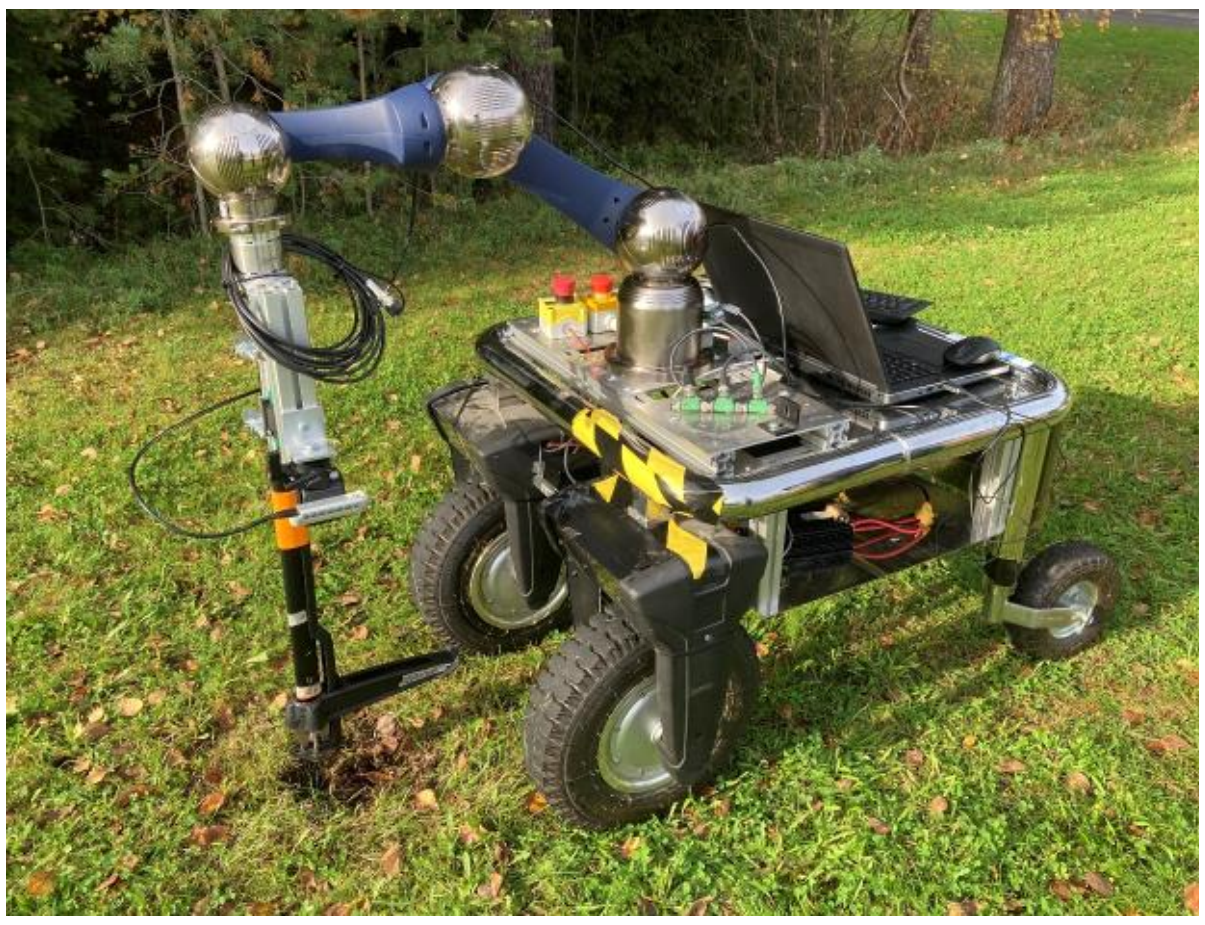

Figure 2. Weeding robot hardware overview.

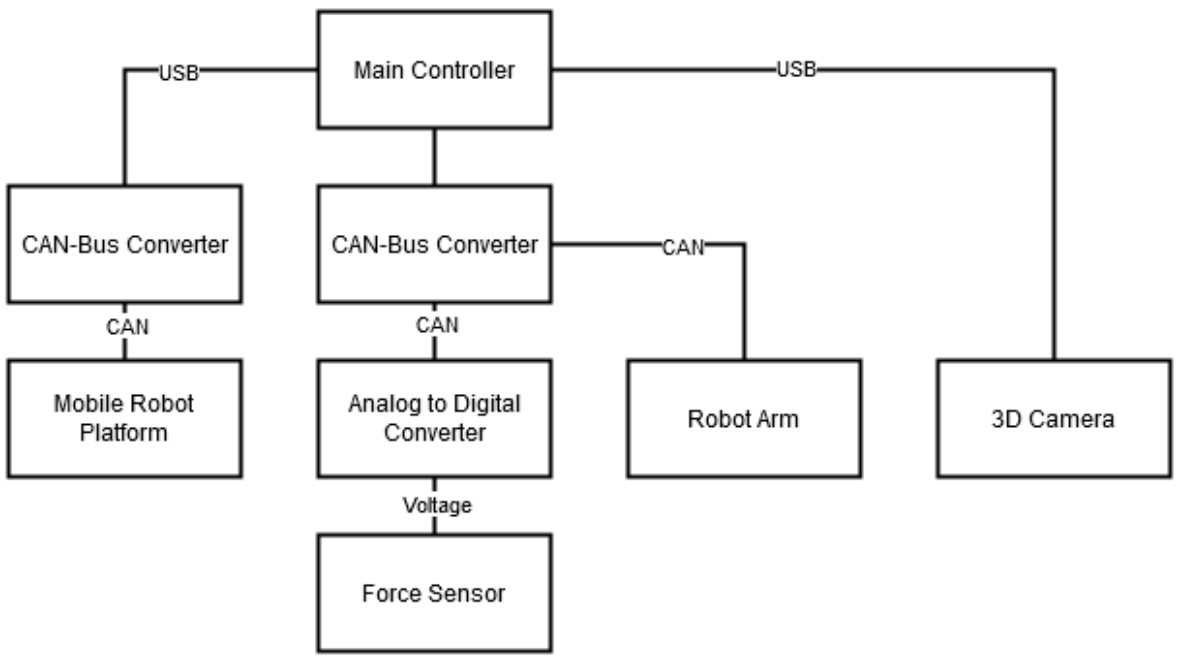

Figure 3. Weeding robot hardware architecture.

The robot is powered by two 48 V lithium-ion batteries in parallel. The voltage is reduced to 24 Volts which is used by all the components. The hardware components are connected to the main controller through the CAN bus for communication. The robot arm and the force sensor are in a separate bus to the motors because they have a different baud rate. The force sensor is connected to an amplifier, which in turn is connected to a CAN-analog input module, where the analog signal is converted to digital CAN data. The electrical schematic of the weeding robot is given in fig 4.

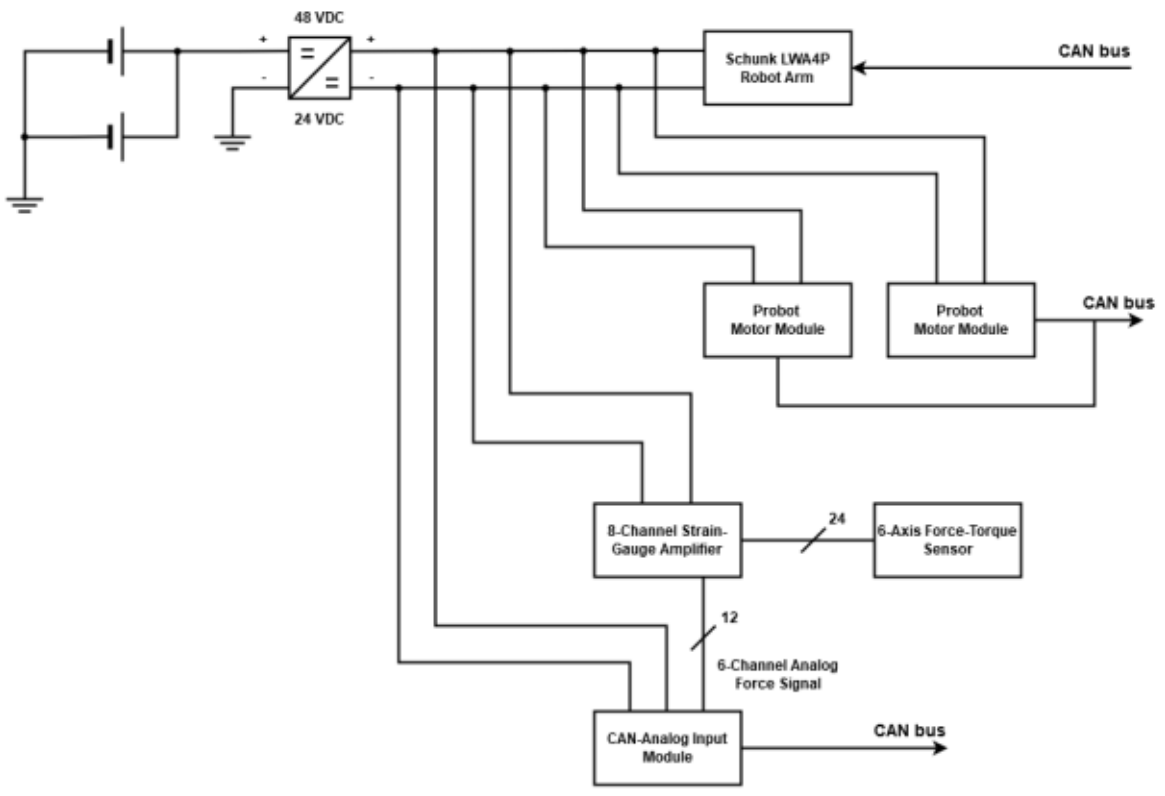

Figure 4. Weeding robot electrical schematic.

The software that controls the robot was written in C++ with Qt. The main program modules (see fig 5 below) are consisting of the weeding robot control software that includes functionality for moving the platform and the robot arm, GNSS based navigation, kinematic model, and control for the Schunk LWA4P robot arm with forward, inverse, and differential kinematics and the planning and executing movement commands with the arm. Also included is the functionality for creating and executing missions for automatic use.

The other pieces of software are the weed detector software that takes 2D and 3D images with an Intel RealSense D415 3D camera and calculates a 3D position of weeds from that image as well as a ROS node for interfacing with the GNSS receiver. The weed detector software uses RealSense SDK to interface with the camera. The control software sends a command to the weed detector to take an image with the camera and the detector responds with the calculated weed position. The U-Blox GNSS receiver ROS node communicates with the receiver itself and posts the received coordinates to a ROS topic. The control software subscribes to that topic to receive the latest coordinates.

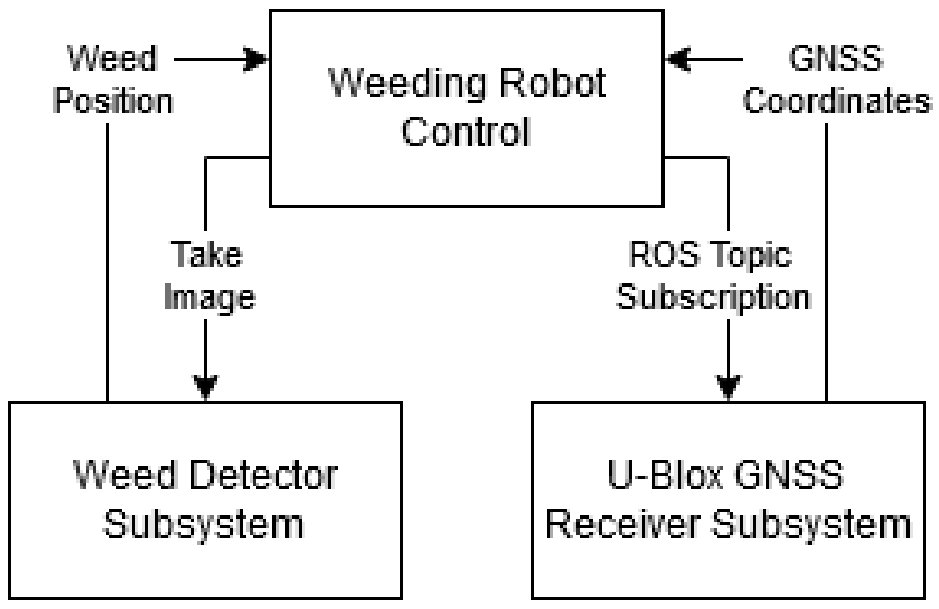

Figure 5. Software subsystems.

The control software functionality is split into multiple classes for easier development and asynchronous execution. The relationships of the classes are illustrated in fig 6, and a short description of each class and its purpose are as follows:

### A. *UI Classes*

- WeedingRobot Control UI is the only UI and parent class that has every other class as a child object, and it also has the UI elements, including control buttons, status messages, and data displays.

### B. *Mission/Task Control Classes*

- MissionControl contains the mission, which is a list of tasks. It sends the next task in the list to TaskControl to execute.
- MissionTaskDialog is a UI class for editing the current mission.
- TaskControl contains information on how each task is executed and it also keeps track of what control mode the robot is in. The modes indicate how the robot is controlled, for example mission mode, joystick mode, and coordinate drive mode.

### C. *Platform Control Classes*

- PathControl is the class that keeps track of the path the platform moves in. It receives coordinate waypoints and creates a path for the platform. It also calculates the direction where the robot needs to go.
- CoordinateDriveDialog is a UI class for selecting coordinates for the robot to move towards.
- TrajectoryControl takes the set direction of the platform and calculates the required motor speeds to turn and move towards that direction.
- MotionControl receives the target motor speeds and modifies them into the format required by the motors. In simulation mode it also sends motion information to LocationEstimator.
- CANBusConnection sends data and receives data to and from the CAN bus.

### D. *Position Sensor Control Classes*

- LocationEstimator receives data from the sensors and uses the EKF to estimate its position and direction.
- SensorData reads data from the IMU sensor and the GNSS receiver and sends it forward.

### E. *Robot Arm Control Classes*

- RobotArmControl controls the robot arm. It has a state machine that contains all the bytes required to be sent in order to enable the arm. It also computes the movement paths for point-to-point motion as well as the movement increments for linear motion. Finally, it has the force control for compliant motion.
- RobotArmMoveDialog is a UI class for sending motion commands to the robot arm.
- RobotKinematics is a class that calculates the forward, inverse, and differential kinematics for the robot arm.
- ForceSensor is a class that converts the voltage signals, from the force sensor via the CAN bus to Newtons and Newton-meters with a calibration matrix. It also filters the force signal with a low-pass Butterworth filter.

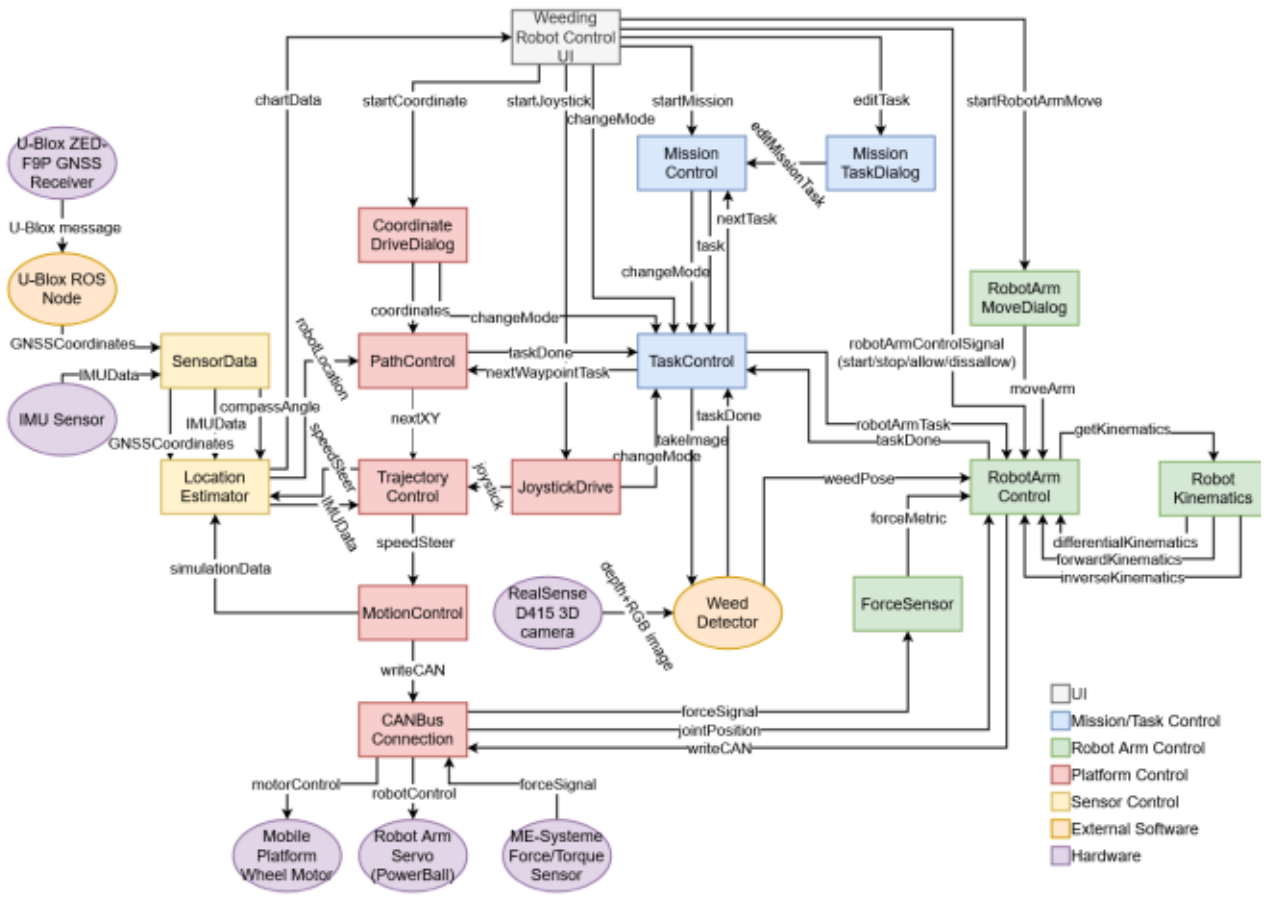

Figure 6. Signal connections with software classes, subsystems, and hardware.

### WEEDING CONTROL

A homogeneous matrix combines the rotation and translation into a single matrix shown in equation 1. Multiplying homogeneous matrices together will give the transformation from one coordinate frame to the other coordinate frame, and the inverse homogeneous transformation matrix gives the transformation backwards in the geometric chain.

$$H = \begin{bmatrix} R_{11} & R_{12} & R_{13} & x \\ R_{21} & R_{22} & R_{23} & y \\ R_{31} & R_{32} & R_{33} & z \\ 0 & 0 & 0 & 1 \end{bmatrix} \quad (1)$$

The robot arm motion is controlled by the coordinate frame (=position and rotation) of its flange (see equation 2). To guide the tool to the weed's location, the weed's position in the robot arm's coordinate frame is determined by the transformations from the base to the flange, through the camera, and to the weed. The inverse transformation from the flange to the tool yields the required flange position. Transformation coordinate frames are shown in figure 7.

$$H = H^{base}_{flange} \times H^{flange}_{camera} \times H^{camera}_{weed} \times {H^{flange}_{tool}}^{-1} \quad (2)$$

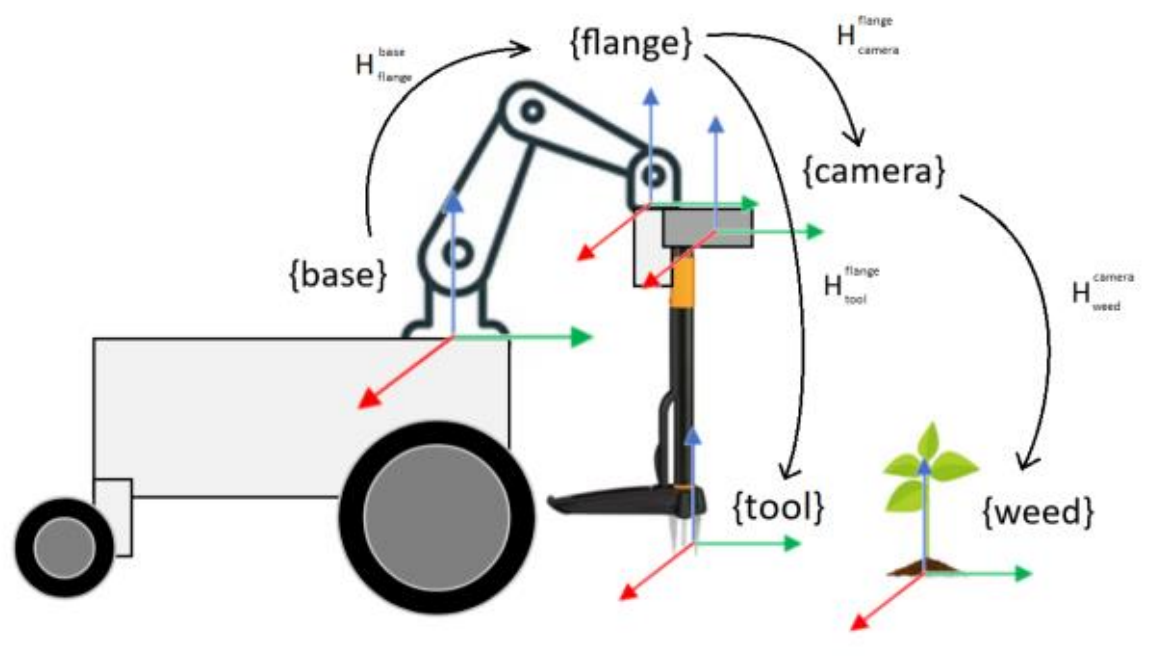

Figure 7. Coordinate frames in the weeding robot’s workspace.

The weed positions in the field are determined by a drone or a tractor and are saved into a weed map. The positions are given in global coordinates that are transformed to field coordinates. Weeding missions are given as task lists where weed locations are given as field coordinates.

The detailed weeding workflow sequence is as follows: First, the robot navigates to the weed position in the field. Second, the robot uses its camera to take a picture of the ground. Third, the weed detection camera system calculates the root position of the weed and projects the pixel coordinate to the depth image. Finally, the 3D position of the weed root is used to calculate the tool path and compliant motion control is applied to ensure the tool remains in contact with the ground. After the weed has been pulled from the ground, the weed is dropped to the ground and the workflow sequence repeats.

### V. PLATFORM CONTROL

The mobile platform control takes care of the movements of weeding robot to the weed locations in the pasture, according to the weed coordinates in the weed map. The navigation system uses the Extended Kalman filter with RTK/GNSS receiver and a compass integration to estimate the position and orientation of the platform [14]. The EKF state vector includes the current estimated position and orientation as well as the velocity of the platform, the input vector has the desired angular velocity and the forward velocity, and the observation vector has the GNSS position and compass orientation. The velocity control of the platform is a simple feed forward scaled value that would cause a cumulative error of the position estimate but with the EKF this error is compensated.

The platform controller calculates the direction it must move to reach the next waypoint. The waypoints are determined by the desired end position and orientation in such a way that there is a direct path segment before the last waypoint to make certain that the platform is pointing in the correct direction. The platform is quite slow to turn, so it is desirable to have a straight path to stabilize its movements. The platform controller calculates the desired angular velocity based on the angle and the distance to the next waypoint in relation to the current position and orientation of the platform. The desired angular velocity is used as the set point and the gyroscope measured angular velocity is used to get the error signal for a PID controller. If the next waypoint is too much to either side of the platform, the speed is lowered so that turning is easier. The wheels of the platform have a very low maximum speed that prevents turning while moving forwards too fast.

### VI. WEED DETECTION

The weed detection works in 2D images by first filtering and thresholding the image to create a black and white image where white pixels correlate to leaves of the plant and black pixels mark the background. The white pixels that are connected are considered as potential segments. The segments need to be clustered and too small and too large segments are discarded. If a segment cannot be reasonably simplified as an ellipse it is also discarded. A principal component analysis is calculated for each leaf segment. The PCA is used to get the centre of the leaf and the largest eigen vector is used to create a line. The lines will gather around the approximate root location of the plant. Intersection points of the lines that are too far from other intersection points are likely outliers and are discarded. A centre point of the remaining intersection points is used as the root location.

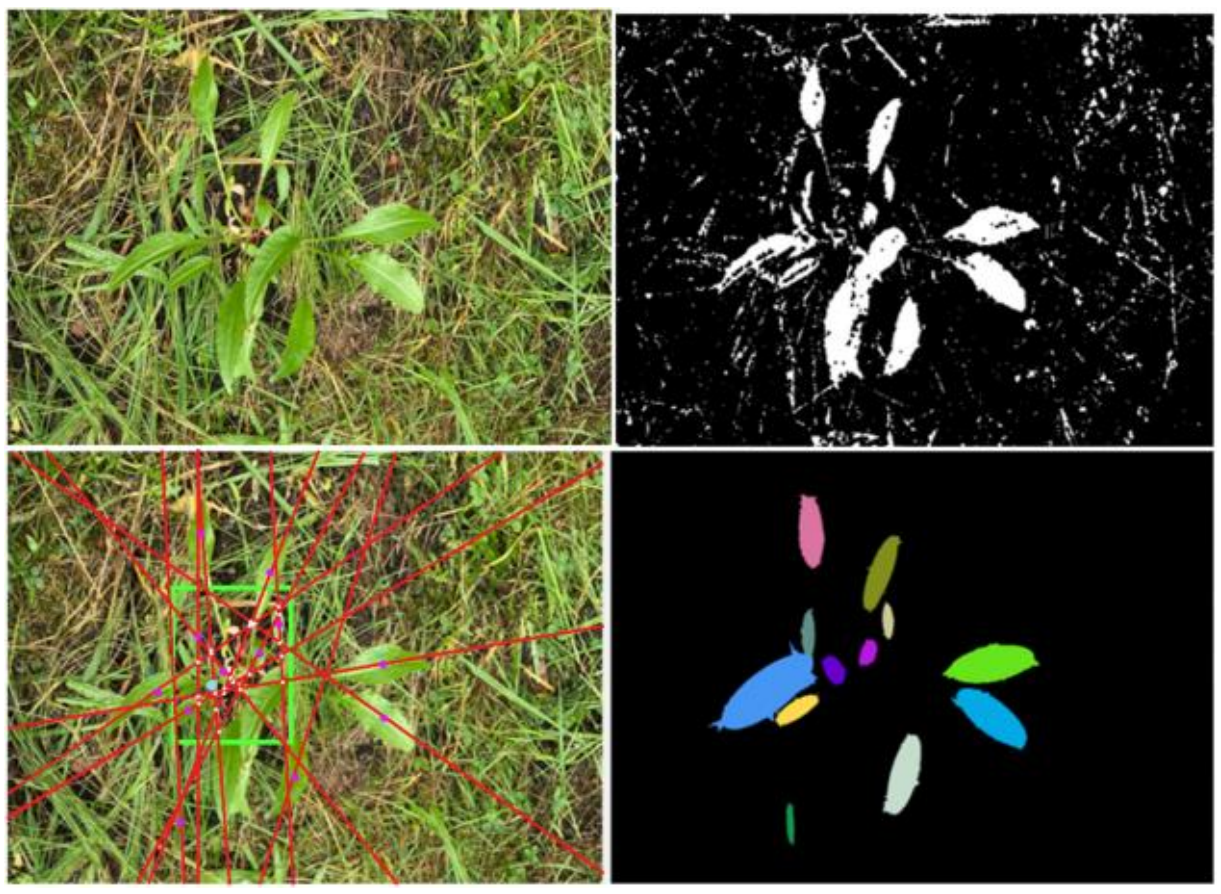

Figure 8 Detection of the weed with some of the processing steps.

In figure 8 the original image, the image processing steps, and the result of the weed detection are shown. The red lines are the directional lines of the segmented leaves, and the intersection points are marked white. The centre of the intersection points is marked light blue. 3D position for this is finally acquired from the corresponding location of the 3D point cloud. Further details of the weed detection can be found from [9].

## VII. MODELING WEEDING ACTION

The weeding motions of the weeding tool were planned by tracking the tool from a video of a person using the tool manually and taking pixel coordinates of certain points of interest in the tool and recording them. The force needed for ground penetration was calculated based on the compression of a spring with a known spring constant. Figures 9 and 10 and graphs in figures 11, 12, and 13 show how the manual weeding action is carried out. In the background there is a scale where each grid element represents 10 cm.

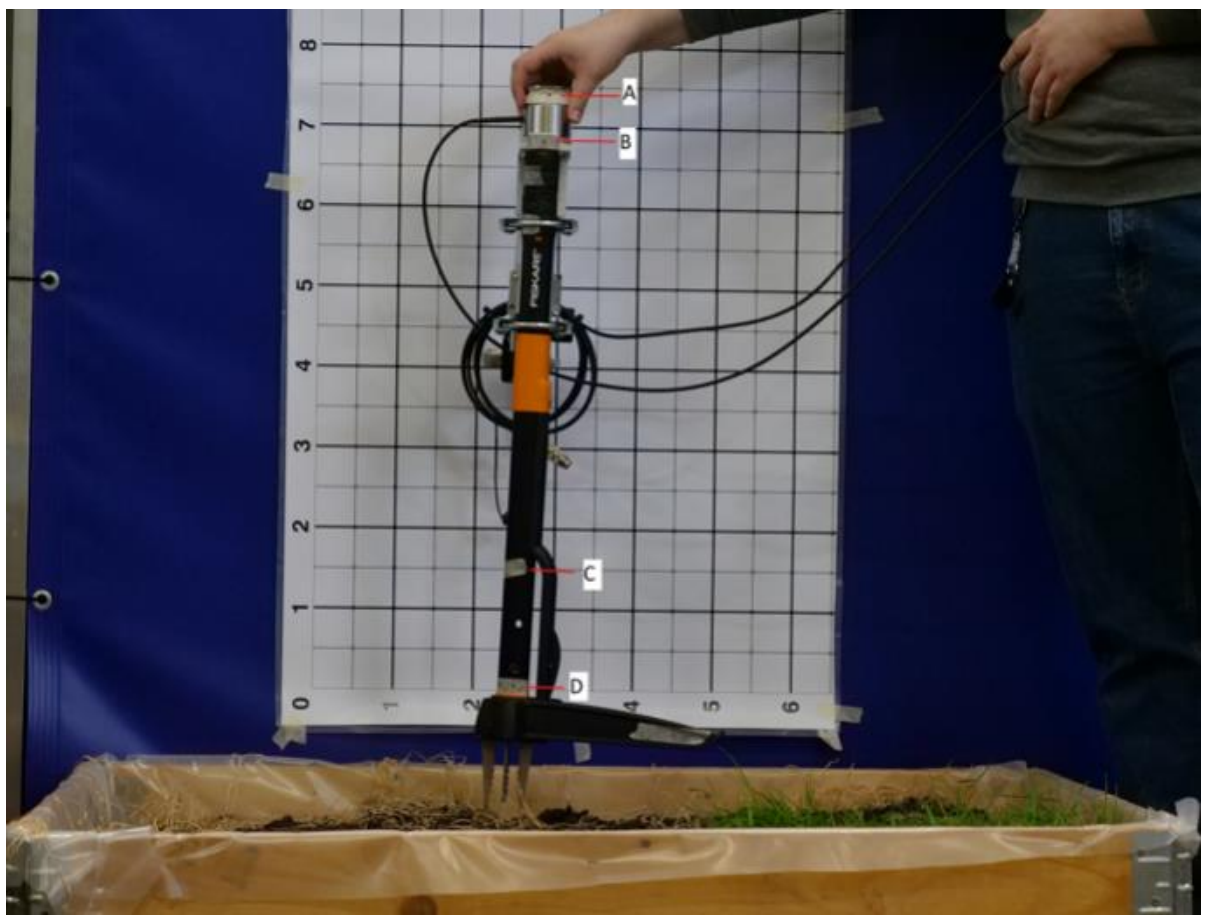

Figure 9. Video frame of the motion planning with the points of interest marked.

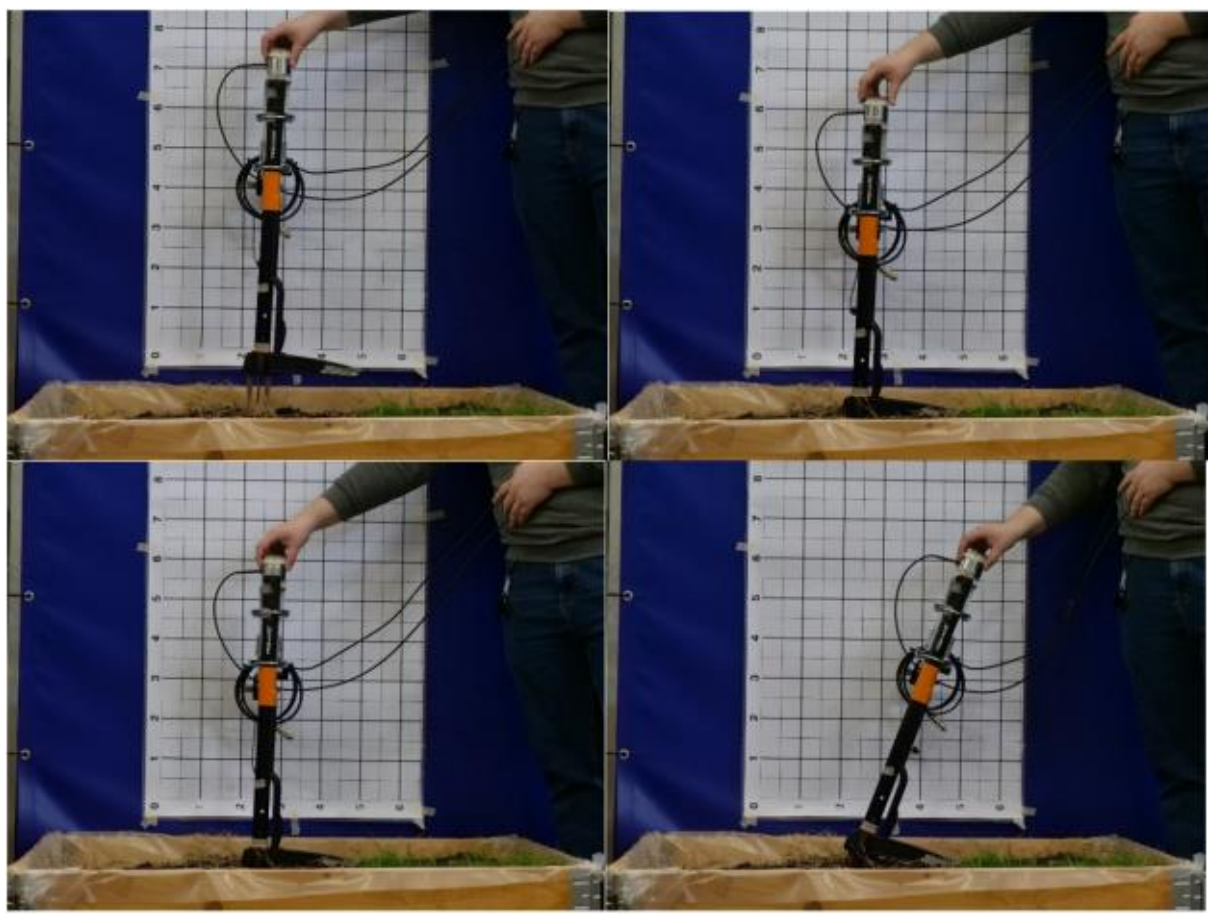

Figure 10. Four frames of the weeding action.

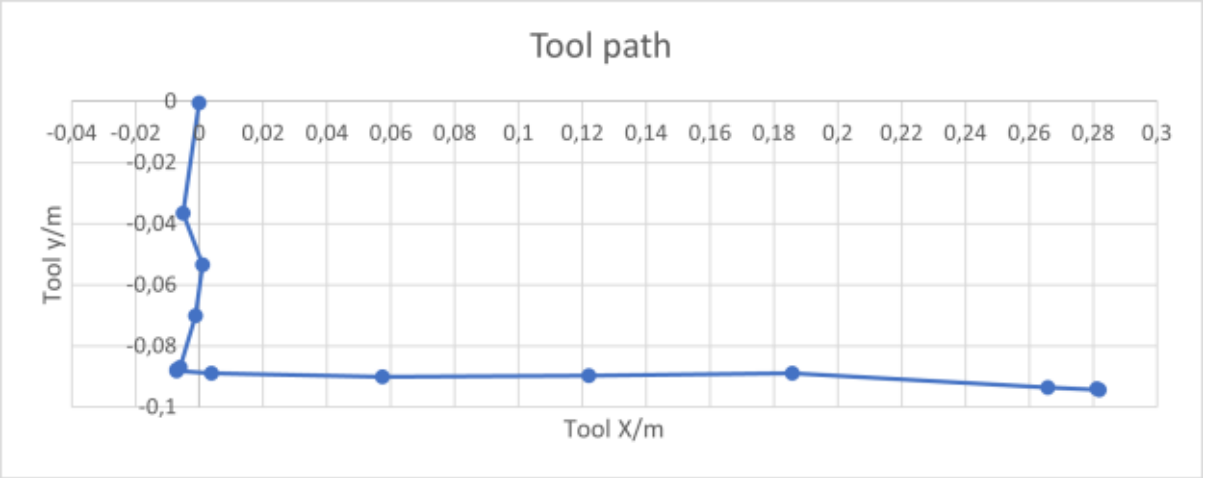

Figure 11. Flange position during the weeding action.

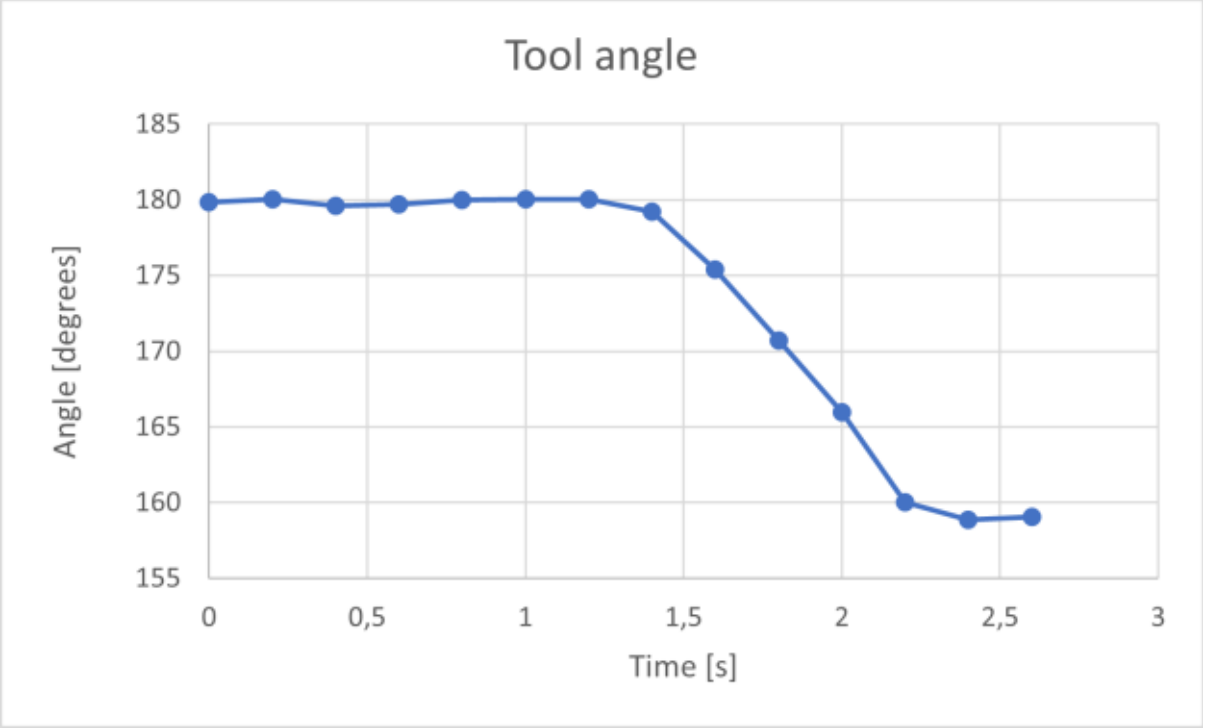

Figure 12. Rotational angle of the flange during the weeding action.

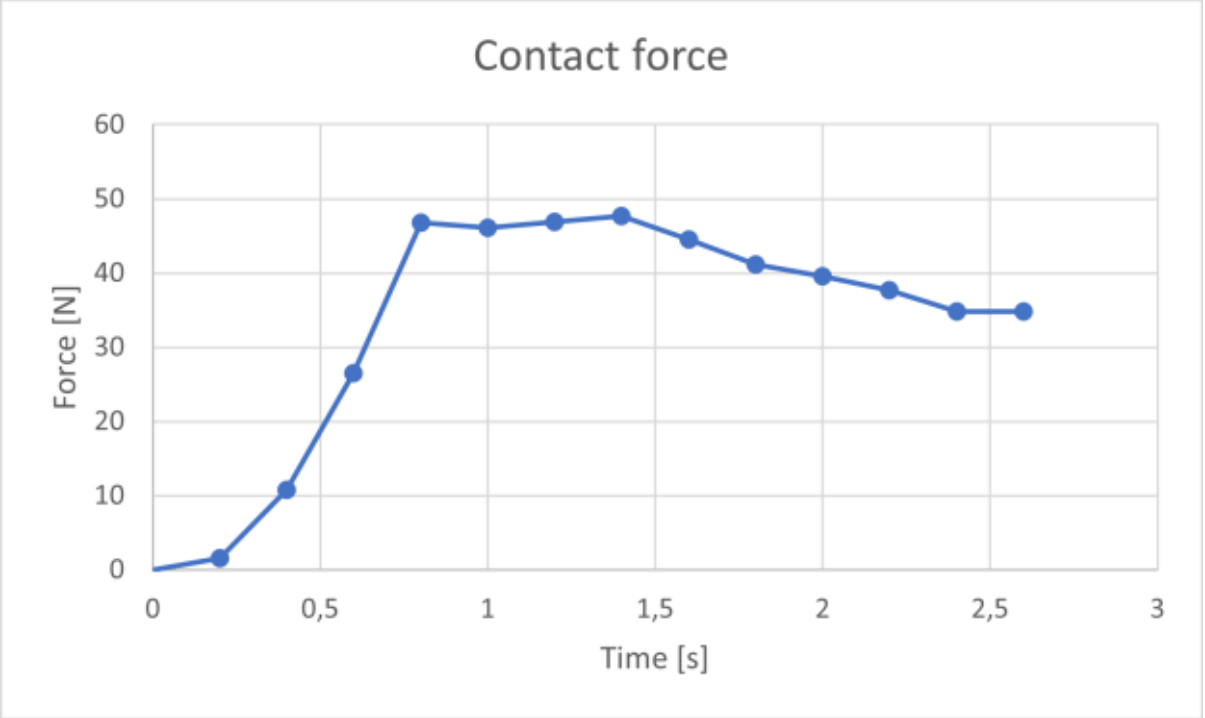

Figure 13. Contact force during the weeding motion.

## VIII. COMPLIANT CONTROL FOR THE ROBOT ARM

The force control is used to maintain the contact force of the weeding tool with the ground. This is done to ensure that the weeding tool uses the lever action properly, because the motion of the fingers of the tool are mechanically synchronized to the lever motions. Compliant motions are split into two motions: guarded and unguarded motions. In the guarded motion, the arm is lowered towards the ground at a constant velocity, until a specified force is detected. The distance lowered is saved as $\Delta X_{guard}$. In the unguarded motion the set motion path of the weeding action is followed with the addition of $\Delta X_{guard}$.

The contact force is maintained by a proportional control, where the desired force is the set point, and a force signal is used to get the error signal for the controller. The measured force is filtered using a low pass 3rd order Butterworth filter with a cut-off frequency of 10 Hz. The output of the P-controller is another position increment $\Delta X_f$ which is added to the modified path $\Delta X_{path}$. $\Delta X_f$ is in the direction of the z-axis of the tool, which is also the axis where the force is measured. The modified and force-controlled path is fed into the inverse differential kinematics to get the next motion increment in joint space. The control principle is described as a block diagram in figure 14.

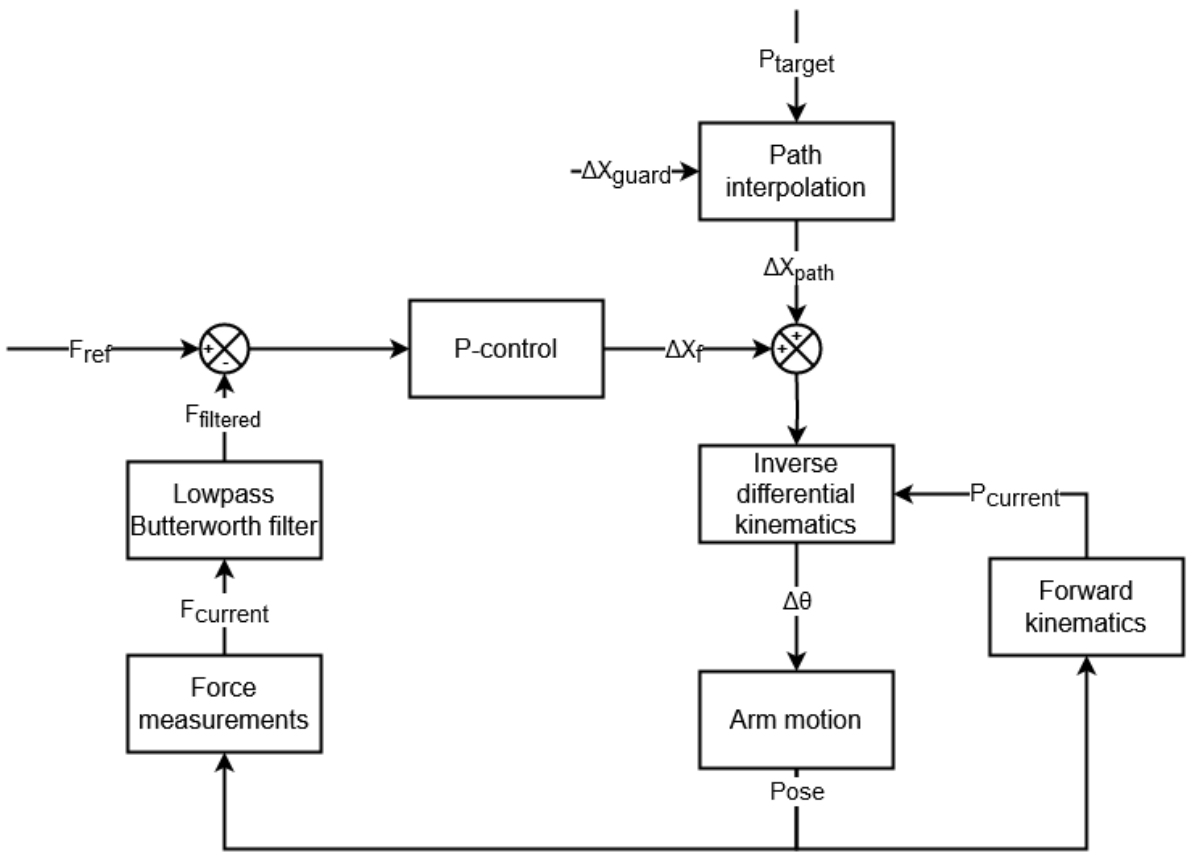

Figure 14. Flow chart of the force control system.

## IX. EXPERIMENTAL TESTS AND EVALUATION

Tests were done outdoors in a grass field shown in figures 16 to 18. The workflow test sequence was as follows: First, the robot navigated to the weed position in the field. Second, the robot used its camera to take a picture of the ground. Third, the machine vision system calculated the root position of the weed and projected the pixel coordinate to the depth image, from which the 3D position was calculated. Finally, the 3D position was used to calculate the tool path and compliant motion control was used during the process to ensure the tool remained in contact with the ground. After the weed had been pulled from the ground, the workflow sequence repeated.

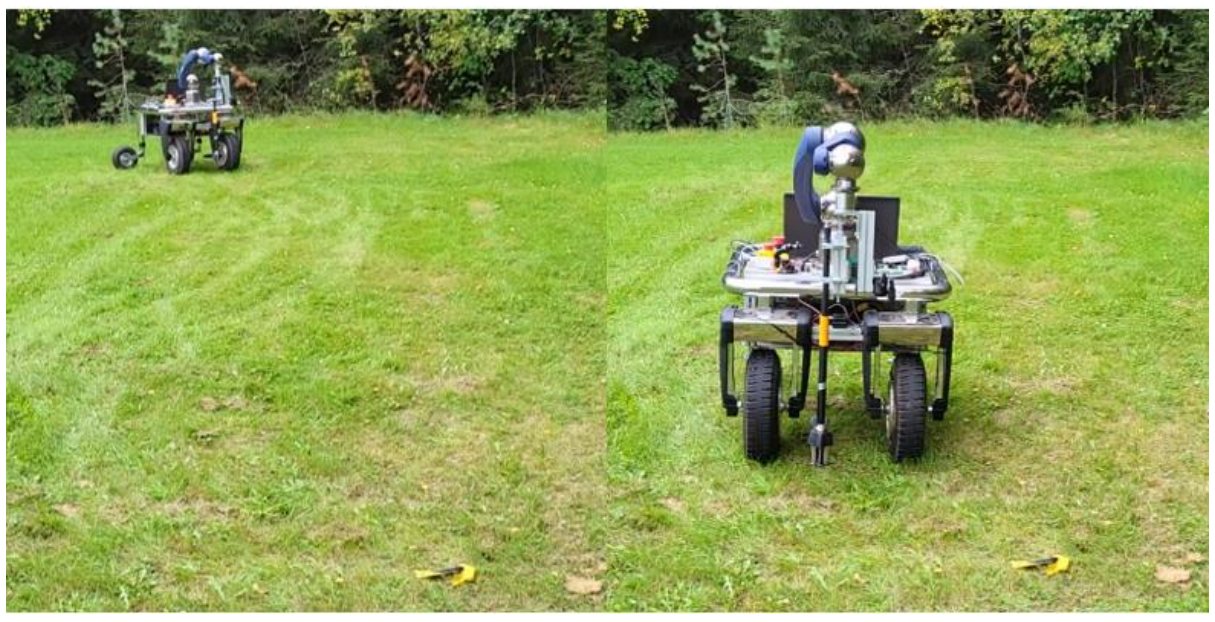

Figure 15. Weeding robot navigating autonomously in the field.

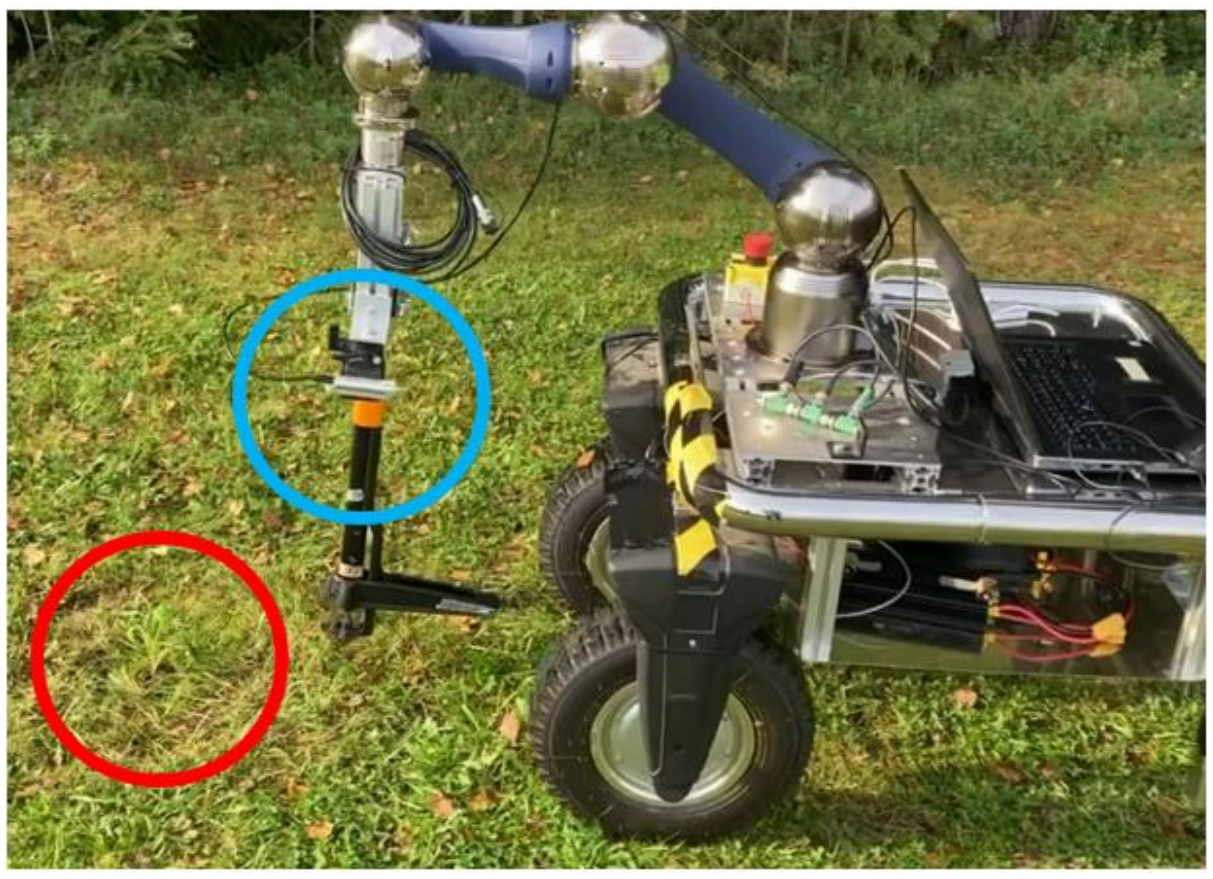

Figure 16. Camera taking a picture of the weed. Blue is the camera, red is the weed.

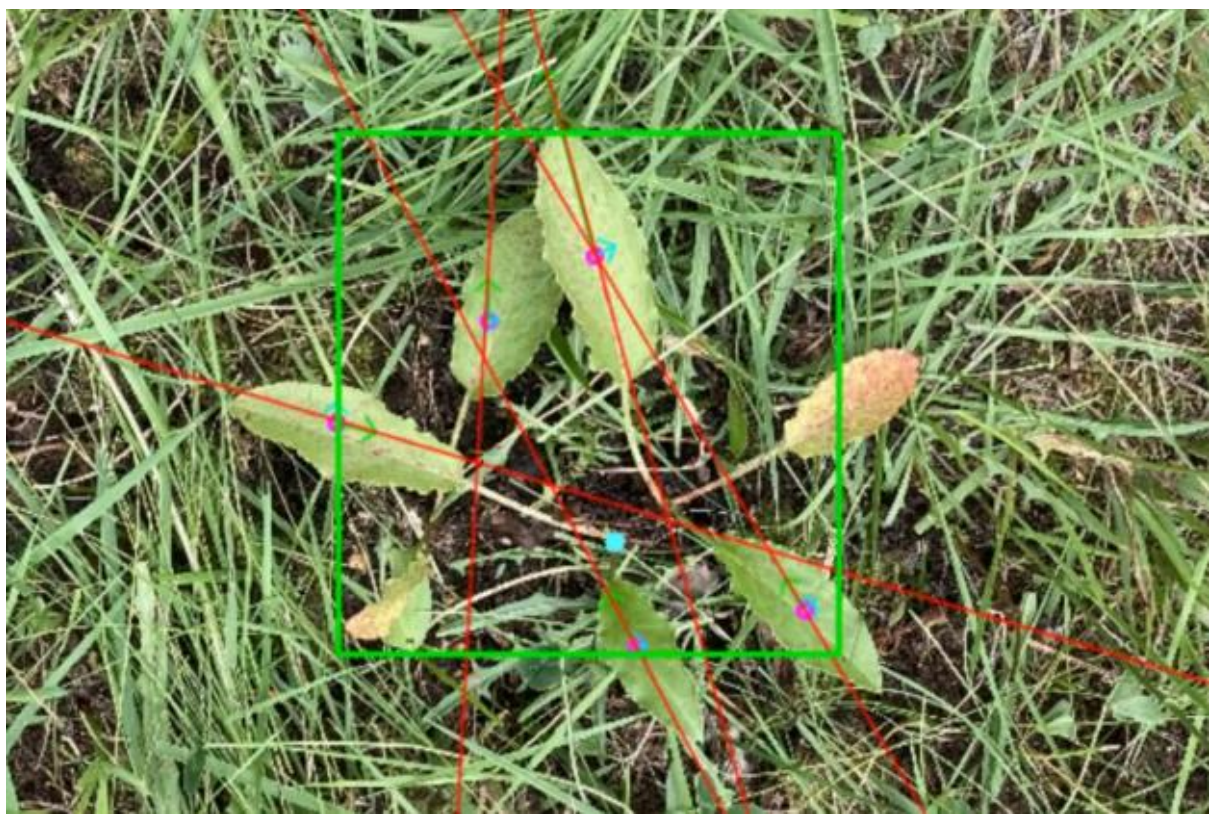

Figure 17. Weed root position detected.

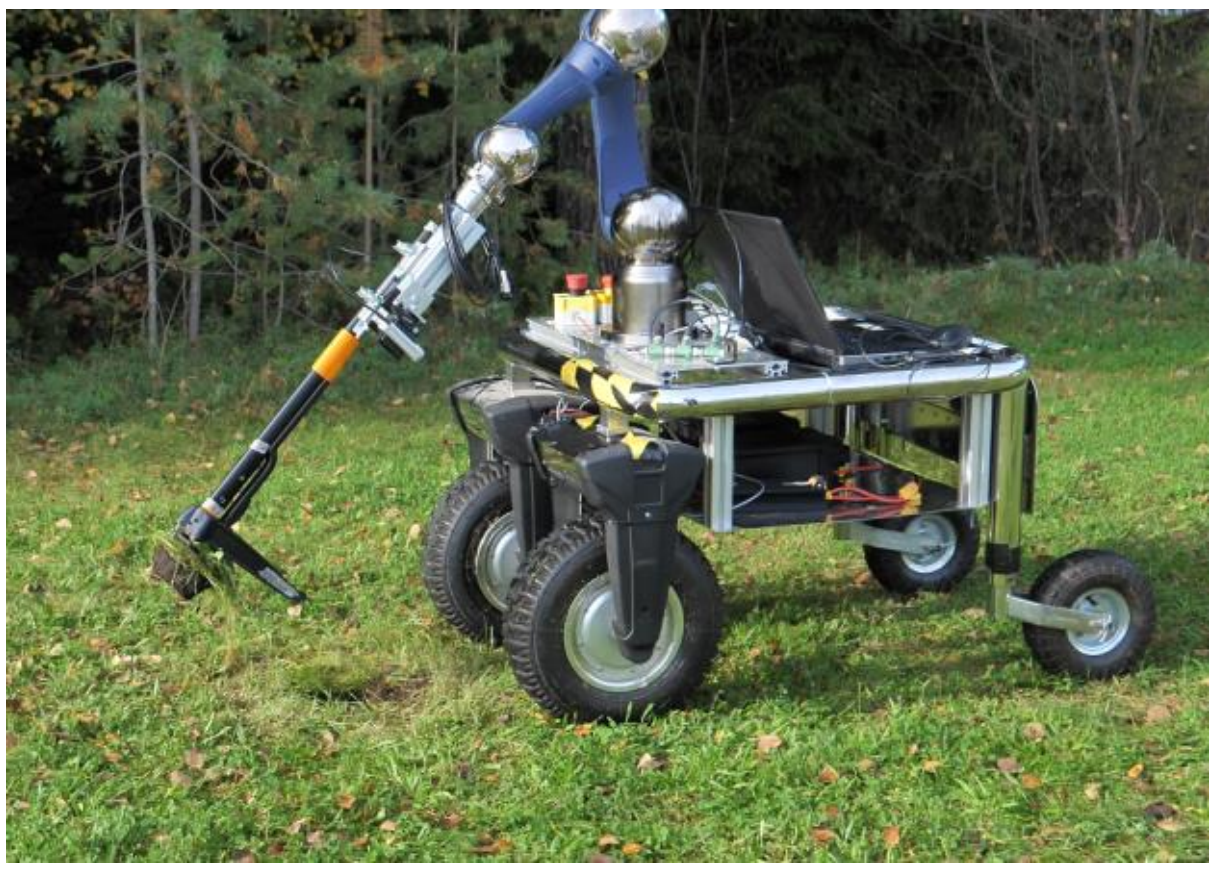

Figure 18. Weed picked from the ground.

In general, when a weed was detected, the weeding action was reliably executed, and the weed was removed.

RTK/GNSS navigation based on U-blox virtual base station worked well to find the weed location accurately. However, the weed detection suffered sometimes from unrelated ground clutter and image overexposure because of strong sunlight. Also, the robot arm did not have enough strength to penetrate the tool to the ground in harder ground areas.

All in all, the tests showed the feasibility of using consumer grade camera technology and weeding tool as a basis for automatic mechanical weeding. The details of the technical validation tests are given in table 1.

TABLE I. TECHNICAL VALIDATION PARAMETERS, TARGETS AND ACHIVED VALUES

| **Validation Parameter** | **Target Value** | **Achieved Value** | **Comment** |
|---|---|---|---|
| Accuracy of the mobile platform to move to a given location | ±0.3 m | ±0.25 m | Based on approximate visual observations; covers the RTK/GNSS navigation and odometry accuracy |
| Accuracy of the detected weed location in the mobile platform coordinate frame | ±0.02m | ±0.02 m | Locating accuracy of the root of the seedling when it has been detected |
| Accuracy of moving the weeding tool to the weed root | ±0.01m | ±0.005 m | Covers the hand-eye and tool calibration inaccuracies |
| Weed removal capability versus the needed force | 50 - 150 N ± 1 N | 50N ± 1N | Force towards the ground, covers the torque around the lever end point of the weeding tool; with loose soil 50 N is enough |

## X. CONCLUSIONS

Automatic weeding technologies have attained a lot of attention lately, because of the harms and challenges weeds are causing for livestock farming. We have presented an automatic mechanical weeding robot that removes weeds from pastures without the use of herbicides. It is based on a lightweight mobile platform, lightweight robot arm with a force/torque sensor, consumer grade 3D camera, and weeding tool. Tests in a real pasture have shown the feasibility of these technologies for robotic weeding.

## ACKNOWLEDGMENT

This work was part of the FlexiGrobots project, funded by the EU's Horizon 2020 research and innovation programme (grant agreement No 101017111) and VTT Technical Research Centre of Finland Ltd., which is greatly acknowledged by the authors.

## REFERENCES

[1] Stewart-Wade, Sally & Neumann, S & L. Collins, L & Boland, Greg. (2002). The biology of Canadian weeds. 117. Taraxacum officinale G. H. Weber ex Wiggers. Canadian Journal of Plant Science. 82. 825-853. 10.4141/P01-010.

[2] S.Yorke, Oxalate poisoning. In: http://www.flockandherd.net.au/edition/poisonousplants2005/oxalate-poisoning.html (18.03.2024)

[3] I. Babiker, W. -F. Xie and G. Chen, "Recognition of Dandelion Weed via Computer Vision for a Weed Removal Robot," 2019 1st International Conference on Industrial Artificial Intelligence (IAI), 2019, pp. 1-6, doi: 10.1109/ICIAI.2019.8850795.

[4] Mohanty S P, Hughes D P and Salathé M (2016), Using Deep Learning for Image-Based Plant Disease Detection. Front. Plant Sci. 7:1419. doi: 10.3389/fpls.2016.01419

[5] X. Sun, G. Li, P. Qu et al. Research on plant disease identification based on CNN, Cognitive Robotics 2 (2022) 155–163

[6] Jana Wäldchen, Patrick Mäder, Plant Species Identification Using Computer Vision Techniques: A Systematic Literature Review. Arch Computat Methods Eng (2018) 25:507–543, https://doi.org/10.1007/s11831-016-9206-z

[7] Erin Winick, Weed-killing robots are threatening giant chemical companies' business models. May 22, 2018. In: https://www.technologyreview.com/2018/05/22/142786/weed-killing-robots-are-threatening-giant-chemical-companies-business-models/

[8] Rene Koerhuis, Odd.Bot robot takes on herbicide free weed elimination. Future Farming, December 20, 2018. In: https://www.futurefarming.com/Tools-data/Articles/2018/12/OddBot-robot-takes-on-herbicide-free-weed-elimination-375027E

[9] Känsäkoski N., Heikkilä T., Kotaniemi J, Detection and Localizing of Rumex Seedlings for Robotic Weeding. IEEE/ASME MESA 2022 – 18th Int. Conference on Mechatronic, Embedded Systems and Applications, 28-30 November 2022, Hybrid Event - Taipei (Taiwan). 7 p.

[10] Leonardo Enrique Solaque Guzmán et. al., Weed-removal system based on artificial vision and movement planning by A* and RRT techniques. Acta Sci., Agron. 41, 2019. https://doi.org/10.4025/actasciagron.v41i1.42687

[11] R. Aravind, M. Daman and B. S. Kariyappa, "Design and development of automatic weed detection and smart herbicide sprayer robot," 2015 IEEE Recent Advances in Intelligent Computational Systems (RAICS), 2015, pp. 257-261, doi: 10.1109/RAICS.2015.7488424

[12] Xin Jin, Lumei Tang, Ruoshi Li, Bo Zhao, Jiangtao Ji, Yidong Ma, Edge recognition and reduced transplantation loss of leafy vegetable seedlings with Intel RealsSense D415 depth camera, Computers and Electronics in Agriculture, Volume 198,2022, 107030, ISSN0168-1699, https://doi.org/10.1016/j.compag.2022.107030.

[13] Xin Jin, Ruoshi Li, Qing Tang, Jun Wu, Lan Jiang, Chongyou Wu, Lowdamage transplanting method for leafy vegetable seedlings based on machine vision, Biosystems Engineering, Volume 220, 2022, Pages 159-171, ISSN 1537-5110, https://doi.org/10.1016/j.biosystemseng.2022.05.017.

[14] Atsushi Sakai, Extended Kalman Filter Localization. In: https://atsushisakai.github.io/PythonRobotics/modules/localization/extended_kalman_filter_localization_files/extended_kalman_filter_localization.html (19.03.2024)

[15] Custom Mobile Robot Solutions. In: https://probot.fi/en/custom-mobile-robot-solutions (20.03.2024)